\pdfoutput=1
\documentclass[11pt]{article}

\usepackage{ACL2023}
\usepackage{times}
\usepackage{latexsym}

\usepackage[T1]{fontenc}
\usepackage[utf8]{inputenc}

\usepackage{microtype}

\usepackage{inconsolata}

\usepackage{colortbl}
\usepackage{color}
\usepackage{helvet}
\usepackage{courier}
\usepackage{natbib}  
\usepackage{caption}
\usepackage{graphicx}
\usepackage{algorithm}
\usepackage{algorithmic}
\usepackage{bm}
\usepackage{amsmath}
\usepackage{amssymb}

\newcommand{\ie}{\textit{i}.\textit{e}., }
\newcommand{\eg}{\textit{e}.\textit{g}., }

\usepackage{booktabs}
\usepackage{multirow}

\usepackage{newfloat}
\usepackage{listings}
\DeclareCaptionStyle{ruled}{labelfont=normalfont,labelsep=colon,strut=off} % DO NOT CHANGE THIS
\floatstyle{ruled}
\newfloat{listing}{tb}{lst}{}
\floatname{listing}{Listing}

\title{Transforming Visual Scene Graphs to Image Captions}

\author{\\Xu Yang$^{1}$ \quad Jiawei Peng$^1$ \quad Zihua Wang$^1$\quad Haiyang Xu$^2$\footnotemark[1]  \quad Qinghao Ye $^2$\\
\quad Chenliang Li$^2$ \quad Songfang Huang$^2$ \quad  Fei Huang$^2$ \quad  Zhangzikang Li$^1$ \quad Yu Zhang$^1$\footnotemark[1]
\vspace{3pt}\\
\normalsize$^1$ School of Computer Science \& Engineering, Key Laboratory of New Generation Artificial Intelligence\\
\normalsize Technology and Its Interdisciplinary Applications (Southeast University), Ministry of Education, China\\
\normalsize$^2$Alibaba Group\\
\tt\small \{101013120, pengjiawei, zihua, zhang\_yu\}@seu.edu.cn,\{shuofeng.xhy, yeqinghao.yqh, \\
\tt\small lcl193798, songfang.hsf\}@alibaba-inc.com, \{feirhuang, lizhangzikang\}@gmail.com
}

\begin{document}

\setlength{\abovedisplayskip}{2pt}
\setlength{\belowdisplayskip}{2pt}
\maketitle

\renewcommand{\thefootnote}{\fnsymbol{footnote}}
\footnotetext[1]{Corresponding authors.}

\begin{abstract}
We propose to TransForm Scene Graphs into more descriptive Captions (\textbf{TFSGC}). In TFSGC, we apply multi-head attention (MHA) to design the Graph Neural Network (GNN) for embedding scene graphs. 
After embedding, different graph embeddings contain diverse specific knowledge for generating the words with different part-of-speech, \eg object/attribute embedding is good for generating nouns/adjectives. Motivated by this, we design a Mixture-of-Expert (MOE)-based decoder, where each expert is built on MHA, for discriminating the graph embeddings to generate different kinds of words. 
Since both the encoder and decoder are built based on the MHA, as a result, we construct a \textbf{simple and homogeneous} encoder-decoder unlike the previous \textbf{heterogeneous} ones which usually apply Fully-Connected-based GNN and LSTM-based decoder. The homogeneous architecture enables us to unify the training configuration of the whole model instead of specifying different training strategies for diverse sub-networks as in the heterogeneous pipeline, which releases the training difficulty. Extensive experiments on the MS-COCO captioning benchmark validate the effectiveness of our TFSGC. The code is in: \url{https://github.com/GaryJiajia/TSG}.
\end{abstract}

\section{Introduction}
\label{sec:intro}
\begin{figure*}[htb]
  \centering
  \includegraphics[width=15cm]{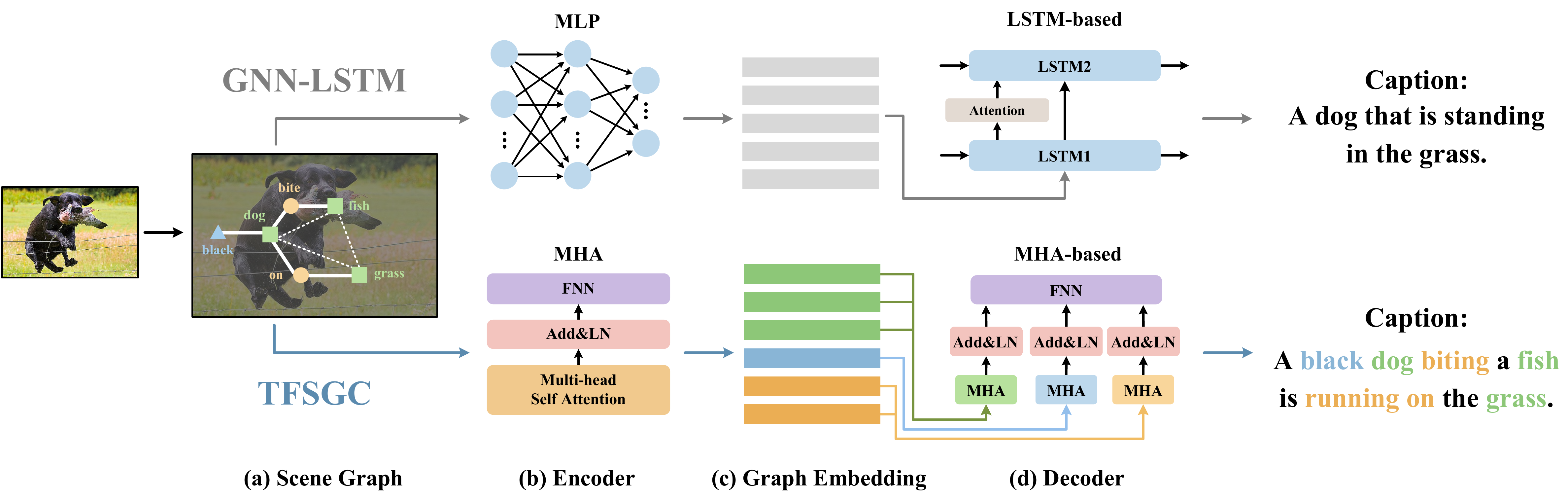}
  \caption{Comparison between traditional heterogeneous GNN-LSTM (top part) and our homogeneous TFSGC model (bottom part). In GNN-LSTM, they use MLP-based GNN and do not discriminate the graph embeddings (grey colour in (c) is used to strengthen such indiscrimination). In TFSGC, we use MHA to design the GNN and the decoder and discriminate diverse graph embeddings (different colours in (c) are used to strengthen such discrimination).}
  \label{fig:overall}
  % \vspace{-0.2in}
\end{figure*}
Image captioning, which aims to generate one sentence for describing multi-aspects of an image, has made huge progress since the proposal of the encoder-decoder framework~\cite{vinyals2015show,xu2015show}. Such a framework contains one visual encoder to extract a series of visual features from the image and one language decoder to generate captions from the extracted visual features. Since the visual encoder is usually well pre-trained by image classification and object detection, the extracted features contain abundant knowledge of the object categories, which enables the captioning model to generate object-abundant captions. 

However, object category is not the only visual pattern that matters for high-quality captions~\cite{anderson2018bottom,jiang2020defense}, object attributes and relations also play significant roles in generating descriptive captions, \ie the caption containing multi-aspects of an image. Motivated by this, researchers propose to incorporate additional semantic knowledge, \eg object categories, attributes, and relations, into the captioning models by using the scene graph as the mediator~\cite{yao2018exploring,yang2020auto}. Scene graphs assign each object node with certain attribute nodes and some pairwise objects with certain relation nodes. These nodes are represented by the corresponding semantic tags, \eg as shown in Fig.~\ref{fig:overall}, the object ``dog'' is assigned with the attribute ``black'' and the pairwise objects ``dog'' and ``fish'' have the relation ``bite'' in between. To exploit the scene graph, Graph Neural Network (GNN)~\cite{battaglia2018relational} is deployed to embed the graphs and the output embeddings are input to the decoder for captioning. 

The top part of Fig.~\ref{fig:overall} shows the pipeline of the previous popular GNN-based captioning model~\cite{yao2018exploring,yang2020auto}, which implements GNN as a few Fully-Connected (FC) and non-linear activation layers. To update the node embedding, this GNN maps the concatenated neighbour embeddings into the new one~\cite{xu2018how}. Then the updated graph embeddings are input into the language decoder that contains a few LSTM layers and an attention module. The LSTM layers are used to generate the context vector based on the partially generated captions. This context vector works as the query in the attention module for determining which graph embeddings should be used to generate the next word. Compared with the models without GNN, this GNN-LSTM pipeline usually gets better performance.

However, this GNN-LSTM framework implies two flaws which hinder the further improvement of applying scene graphs. First, FC-based GNN and LSTM do not share the same building blocks and thus the constructed model is a \textbf{heterogeneous} structure, which requires well-chosen training strategies, \eg choosing different learning rates or optimizers for different sub-networks, to achieve the best performance~\cite{yang2020auto}. Finding such training configurations is a labour-intensive process. Second, the graph embeddings are indiscriminately selected during captioning (the grey embeddings in Fig.~\ref{fig:overall} top (c) denote such indiscrimination), which causes less descriptive captions. While intuitively, different kinds of node embeddings should be used for generating the words with diverse part-of-speech (POS), \eg the object/attribute/relation embeddings should be more responsible for the nouns/adjectives/verbs, respectively~\cite{yang2019learning}.

To alleviate the above-mentioned flaws, we propose a novel \textbf{homogeneous} captioning model to Transform Scene Graphs (\textbf{TFSGC}) into captions. Our TFSGC is built based on the Transformer~\cite{vaswani2017attention} since 
it is more powerful than LSTM in image captioning~\cite{herdade2019image, Li_2019_ICCV,cornia2020meshed}. TFSGC is homogeneous since we use multi-head attention (MHA) to design both the graph encoder to embed the scene graphs and the language decoder to generate the caption. 

Our design principle is quite simple where we do not need to revise the self-attention operation but only need to reformulate the input data structure. Specifically, to design GNN by MHA, we first linearize the scene graph into a token sequence and introduce a binary mask to index which two nodes are connected in the graph. Then we use the masked MHA operation to deal with this linearized token sequence for graph embedding. In this process, each token embedding is added by a learnable type embedding to index the token type (\eg object/attribute/relation) and we will show that such type embedding can help distinguish the edge type during the attention calculation.

After graph operation, we get a series of object/attribute/relation embeddings, which will be used in the decoder for captioning. To make the decoder discriminate different embeddings for generating different words, we learn from MOE networks~\cite{jacobs1991adaptive,xue2022go,du2022glam} to revise the original Transformer decoder with two strategies. First, as Fig.~\ref{fig:overall} bottom (d) shows, we use three encoder-decoder attention layers, which are built on MHA, as three experts to address object/attribute/relation embeddings, respectively. Second, we incorporate an attention-based soft routing network to discriminate which kinds of embeddings should be more responsible for generating the next word. Both the MOE-decoder and the type embedding in the encoder help distinguish node embeddings for better captions. We carry exhaustive ablation studies and comparisons to validate the effectiveness of TFSGC and it achieves 132.3/138.6/139.5 CIDEr scores when using BUTD/Patch/VinVL features.

\section{Related Work}
\noindent\textbf{Image Captioning.}
For a long time, the attention-based CNN-LSTM pipeline~\cite{vinyals2015show,xu2015show} is the most popular backbone for captioning and various techniques have been added into it for better performance, including building stronger visual encoders~\cite{lu2018neural,jiang2020defense}, designing more advanced attention mechanisms~\cite{anderson2018bottom,wang2020show}, incorporating semantic knowledge~\cite{you2016image,gan2017semantic,yao2018exploring,yang2020auto}, and exploiting language structure~\cite{lu2017knowing,yang2019learning}. 

Recently, Transformer~\cite{vaswani2017attention} has gradually substituted LSTM as the mainstream language decoder in image captioning~\cite{herdade2019image, Li_2019_ICCV} since it achieves better performances than the LSTM-based models. Based on this new backbone, researchers develop more advanced strategies for further improving the effectiveness, including designing more sophisticated attention mechanisms~\cite{huang2019attention,pan2020x}, introducing additional memory blocks~\cite{cornia2020meshed,yang2021causal}, distilling knowledge from the large-scale pre-training models~\cite{radford2021learning,li2021align, xu2021e2e}, and exploiting Transformer-based visual encoders~\cite{2022End,fang2022injecting}, modularized design for large-scale multi-modal pretraining~\cite{li2022mplug,xu2023mplug,ye2023mplug}. Since the recently proposed SOTA models use Transformer as the backbone, we also built TFSGC based on Transformer for fair comparison.

\noindent\textbf{Graph Neural Network (GNN).}
Scene Graph abstracts the major visual patterns in a visual scene as a graph. It is usually used as the mediator to narrow the gap between the vision and the language domains. To incorporate scene graphs into deep networks, GNN~\cite{battaglia2018relational} is used to embed the discrete node labels into dense embeddings. However, most of the previous GNNs are MLP-based~\cite{yang2020auto,yao2018exploring,xu2018how,milewski2020scene,zhong2020comprehensive}, which may limit the effectiveness of embedding scene graphs in a Transformer architecture. In our research, we design an MHA-based GNN to remedy this limitation. Moreover, noisy scene graphs may damage the performances~\cite{nguyen2021defense}, so we use better scene graph parsers to minimize the impact of noise on our model.

\noindent\textbf{Mixture of Experts (MOE).} The major idea of MOE is to construct a network with lots of experts where different experts deal with diverse samples~\cite{jacobs1991adaptive,shazeer2017outrageously}. When a sample is input to the MOE network, a routing network will decide which experts should be more responsible for this input. Thus MOE naturally fits our case where we hope diverse experts can discriminate graph embeddings for generating the words with different POS. Different from the existent MOE-based Transformer~\cite{lepikhin2020gshard,xue2022go,du2022glam} which applies various feed-forward networks as different experts, we set three encoder-decoder attention layers as different experts where the query is set to the same context vector while the key and value are set to object/attribute/relation embeddings.

\section{Revisiting of Transformer}
\label{sec:revisit}
We first revisit the Transformer-based captioning model and then introduce how to revise it to get our TFSGC in the next section. Transformer-based model contains a visual encoder to calculate the contexts of the extracted visual features and the output embeddings will be input into a language decoder for captioning. For both the encoder and the decoder, the most elemental building block is the multi-head attention (MHA). Given the query, key, and value matrices: $\bm{Q} \in \mathbb{R}^{N_Q \times d}$, $\bm{K} \in \mathbb{R}^{N_K \times d}$, $\bm{V} \in \mathbb{R}^{N_V \times d}$, MHA calculates the output $\bm{Y}=\textbf{MHA}(\bm{Q},\bm{K},\bm{V})$ as:
\begin{equation} \label{equ:multi-head}
\small
\begin{aligned}
 \textbf{Input:} \quad  &\bm{Q},\bm{K},\bm{V} \\
 \textbf{Att:} \quad  &\bm{A}_i=\text{Softmax}( \frac{\bm{Q}\bm{W}_i^Q(\bm{K}\bm{W}_i^K)^T}{\sqrt{d}} ) \\
 \textbf{Head}:  \quad  &\bm{H}_i=\bm{A}_i\bm{V}\bm{W}_i^V,\\
 \textbf{Multi-Head:} \quad & \bm{H}= [\bm{H}_1,\bm{H}_2,...,\bm{H}_h]\bm{W}^H, \\
 \textbf{Output:} \quad  &\bm{Y}=\text{LN}(\bm{H}+\bm{Q}), \\
\end{aligned}
\end{equation}
where $\bm{W}_i^Q, \bm{W}_i^K, \bm{W}_i^V \in \mathbb{R}^{d \times d_h}$, and $\bm{W}_i^H \in \mathbb{R}^{  d \times d}$ are all trainable matrices; $h$ is the number of attention heads (set to 8 in the experiments) and $d_h=d/h$; $\bm{A}_i$ denotes the $i$-th attention matrix used to calculate the $i$-th head matrix; $[\cdot]$ means the concatenation operation; and LN is the Layer Normalization operation. 

Besides MHA, another important module in Transformer is the Feed-Forward network (FFN): 
\begin{equation} \label{equ:ffn}
\small
    \textbf{FFN}(\bm{Y})=\text{LN}(\text{FC}(\text{ReLU}(\text{FC}(\bm{Y})))+\bm{Y}),
\end{equation}
where FC denotes the fully-connected layer and ReLU denotes the rectified linear function. 

Given MHA and FFN, we can use them to build a Transformer-based captioning model. For the encoder, it stacks 6 identical blocks where each one contains an MHA and an FFN. Given the output of the former block as the input $\bm{X}$, the next block calculates its output as:
\begin{equation} \label{equ:encoder}
\small
\begin{aligned}
 \textbf{Input:} \quad  &\bm{X}, \\
 \textbf{Self-ATT:} \quad  &\bm{Y}=\textbf{MHA}(\bm{Q}=\bm{X},\bm{K}=\bm{X},\bm{V}=\bm{X}), \\
 \textbf{Output}:  \quad  &\bm{Z}=\textbf{FFN}(Y), \\
\end{aligned}
\end{equation}
Note that the variables $\bm{X},\bm{Y},\bm{Z}$ used here are “local variables” for conveniently introducing the work flow of Transformer architecture, whose values will be set to the specific values when introducing the concrete captioning model. In ``Self-ATT'', $\bm{Q},\bm{K},\bm{V}$ are set to the same value and this operation is named as self-attention~\cite{vaswani2017attention}. After stacking 6 blocks defined in Eq.~\eqref{equ:encoder}, a visual encoder is built. For the first block, its input is the extracted visual feature set of the given image. The output of the last block will be input into the language decoder.

For the decoder, it also stacks 6 identical blocks where each one contains two MHAs and an FFN. Given the output of the former decoder block $\bm{X}_D$ and the output of the visual encoder $\bm{X}_E$, the next decoder block calculates its output:
\begin{equation} \label{equ:decoder}
\small
\begin{aligned}
 \textbf{Input:} \quad  &\bm{X}_D,\bm{X}_E \\
 \textbf{Self-ATT:} \quad  &\bm{Y}_1=\textbf{MHA}(\bm{Q}=\bm{X}_D,\bm{K}=\bm{V}=\bm{X}_D), \\
 \textbf{ED-ATT:} \quad  &\bm{Y}_2=\textbf{MHA}(\bm{Q}=\bm{Y}_1,\bm{K}=\bm{V}=\bm{X}_E), \\
 \textbf{Output}:  \quad  &\bm{Z}=\textbf{FFN}(\bm{Y}_2), \\
\end{aligned}
\end{equation}
Note that in ``ED-ATT'', $\bm{Q}$ is set to the output of the former decoder block while $\bm{K},\bm{V}$ are set to the output of the visual encoder, and such operation is called encoder-decoder attention~\cite{vaswani2017attention}. After stacking 6 blocks defined in Eq.~\eqref{equ:decoder}, a language decoder is built. 

For the first block in the decoder, $\bm{X}_D$ in Eq.~\eqref{equ:decoder} is set to the word embedding set of the partially generated captions $\bm{S}=\{s_1, ..., s_t\}$ at the $t$-th time step. For all the decoder blocks, the input $\bm{X}_E$ is set to the same value, which is the output of the visual encoder. The output of the last decoder block $\bm{Z}=\{\bm{z}_1,...,\bm{z}_t\}$ is used to calculate the word distribution of the next word:
\begin{equation}
    P(s_{t+1})=\text{Softmax}(\bm{z}_t).
\end{equation}
Given the ground-truth caption $\bm{S}^{*}$, we can train this model by minimizing the cross-entropy loss:
\begin{equation} \label{equ:equ_celoss}
    L_{XE} = -\log P({S}^*),
\end{equation}
or by maximizing a reinforcement learning (RL) based reward~\cite{rennie2017self}:
\begin{equation} \label{equ:equ_rlloss}
    R_{RL} = \mathbb{E}_{\bm{S}^s \sim P(\bm{S})}(r(\bm{S}^s;\bm{S}^*)),
\end{equation}
where $r$ is a sentence-level metric for the sampled sentence $\bm{S}^s$ and the ground-truth $\bm{S}^*$, \eg the CIDEr-D~\cite{vedantam2015cider} metric.

\begin{figure}[t]
  \centering
  \includegraphics[width=7.5cm]{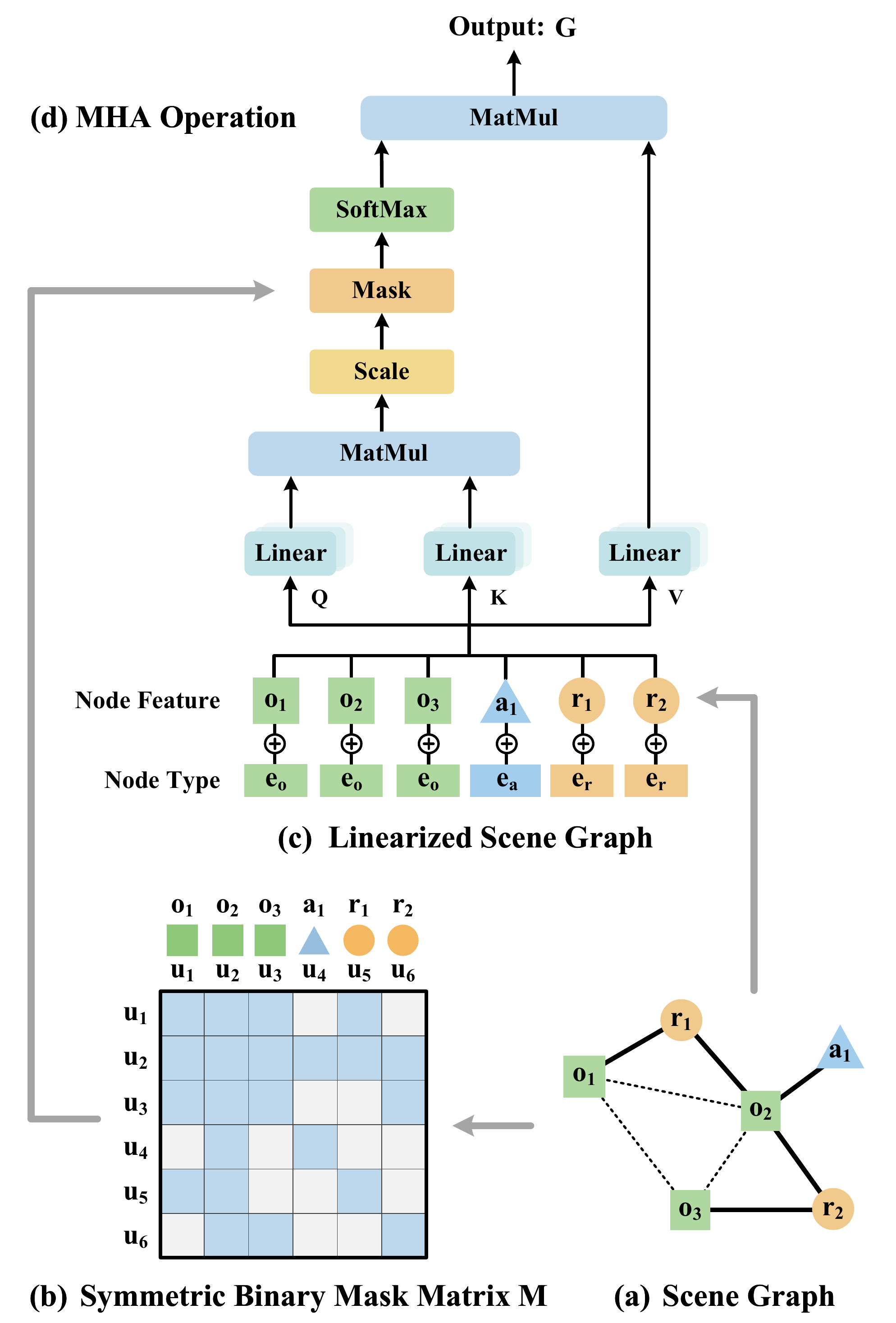}
  \caption{The sketch of the proposed MHA-GNN. In (a), square/triangle/circle demonstrate the object/attribute/relation embeddings, where the dash line means that all the object nodes should be connected for capturing visual contexts. (b) sketches the built binary mask matrix, where the top part shows the original graph embeddings ($\bm{o}/\bm{a}/\bm{r}$). For convenience, we also index the linearized annotation ($\bm{u}$) in the left and top. (c) shows the linearized scene graph, where the top and bottom parts are respectively node feature and type embeddings. (d) details how MHA achieves the graph operation.}
  \label{fig:encoding}
\end{figure}

\section{Transforming Scene Graphs}
\label{sec:TFSGC}
In this section, we introduce how to revise the Transformer to get our TFSGC. We will first show how to get an MHA-based GNN and then introduce how to design an MOE-based decoder.

\subsection{MHA-GNN}
\label{sec:TFSGC_encoder}
A visual scene graph~\cite{krishna2017visual} contains three kinds of node embeddings: object/attribute/relationship embeddings $\bm{o}$/$\bm{a}$/$\bm{r}$. These nodes are connected by the following rules: if an object $\bm{o}_i$ has an attribute $\bm{a}_{k}$, $\bm{o}_i$ and $\bm{a}_{k}$ are connected, \eg $\bm{o}_1$ connects $\bm{a}_{1}$ in Fig.~\ref{fig:encoding} (a). If two objects $\bm{o}_i$ and $\bm{o}_j$ have the relation $\bm{r}_{k}$, we connect $\bm{r}_{k}$ with
$\bm{o}_i$ and $\bm{o}_j$, \eg $\bm{
r}_1$ connects $\bm{o}_1$ and $\bm{o}_2$. Given an image, we extract a series of visual features from the image as the object embeddings: $\{\bm{o}_1,...,\bm{o}_{N_o}\}$. To get the attribute/relation embeddings, we first use the attribute/relation annotations from VG~\cite{krishna2017visual} to train the attribute/relation classifiers to predict the labels. Then we use two learnable embedding layers to respectively transform these labels into the dense attribute/relation embeddings $\{\bm{a}_1,...,\bm{a}_{N_a}\}$/ $\{\bm{r}_1,...,\bm{r}_{N_r}\}$.

Given these original node embeddings, GNN will update each one by aggregating the neighbour embeddings. In the previous GNN-LSTM-based models, GNN is usually deployed by the FC layers, which aggregates the contexts by mapping the concatenated neighbour embeddings to the new one~\cite{yao2018exploring,yang2020auto}. However, in our TFSGC, since Transformer is applied as the backbone, we may more hope to use the basic building block of Transformer to deploy the GNN. Such a design principle has two advantages. First, it alleviates the coding implementation difficulty that we do not need to specify some additional GNN operations. Second, which is more important, when the GNN and the Transformer architecture are homogeneous, the whole model will be more easily trained. For example, we do not need to set different training strategies like the learning rate or optimizer to different sub-networks.

Since MHA (Eq.~\eqref{equ:multi-head}) can learn the context knowledge between the embeddings~\cite{vaswani2017attention}, it can naturally be used to define the graph operation for aggregating the knowledge. We apply the following two steps to do this. Firstly, as shown in Fig.~\ref{fig:encoding}, we linearize the object, attribute, and relation embeddings into one sequence and add the learnable type embeddings as the linearized token set: $\bm{U}=\{ \bm{u}_1,...,\bm{u}_{N} \}$:
\begin{equation}
\small
 \begin{aligned}
 \textbf{Object:} \quad   & \bm{u}_i = \bm{o}_i + \bm{e}_o, & 1 \le i \le  N_o,\\
 \textbf{Attribute:} \quad   & \bm{u}_{N_o+i} = \bm{a}_i + \bm{e}_a, & 1 \le i \le  N_a,\\
 \textbf{Relation:} \quad  & \bm{u}_{N_o+N_a+i} = \bm{o}_i + \bm{e}_r, & 1 \le i \le  N_r,\\
\end{aligned}
\end{equation}
where $\bm{e}_o/\bm{e}_a/\bm{e}_r$ are respectively learnable type embeddings correspond to object/attribute/relation types and $N = N_o+N_a+N_r$. For example, in Fig.~\ref{fig:encoding}, $N_o$/$N_a$/$N_r$ is 3/1/2, the objects $\bm{o_{1:3}}$ become $\bm{u}_{1:3}$, the attributes $\bm{a}_1$ becomes $\bm{u}_{4}$, and the relations $\bm{r}_{1:2}$ become $\bm{u}_{5:6}$.

After linearizing, the topological knowledge of the graph is lost, \ie this token sequence does not show which two nodes are connected or not. To remedy such knowledge, we use a symmetric binary mask matrix $\bm{M} \in \mathbb{R}^{N \times N}$ to all Transformer blocks to mask the attention weights of the unconnected nodes to control whether two nodes are connected or not. If two nodes $\bm{u}_i$ and $\bm{u}_j$ are connected in the original scene graph, we set $M_{i,j}=1$ and $M_{i,j}=0$ otherwise. Specifically, the values of $\bm{M}$ are set as follows: 1) If $\bm{o}_i$ has one attribute $\bm{a}_j$, we set $M_{i,j+N_o}=1$, \eg $\bm{o}_2$ ($\bm{u}_2$) and $\bm{a}_1$ ($\bm{u}_4$) in Fig.~\ref{fig:encoding} are connected and $M_{2,4}=1$. 2) If $\bm{r}_k$ connects with $\bm{o}_i$ and $\bm{o}_j$, we set $M_{i,k+N_o+N_a}=M_{j,k+N_o+N_a}=1$, \eg $\bm{r}_1$ ($\bm{u}_5$) connects $\bm{o}_1$ ($\bm{u}_1$) and $\bm{o}_2$ ($\bm{u}_2$) and $M_{1,5}=M_{2,5}=1$. 3) All the object nodes are connected with each other since they are visual features that their contexts play a key role in captioning~\cite{herdade2019image}. Thus $\forall i,j\leq N_o, M_{i,j}=1$. 4) Since the scene graph is an undirected graph, $\bm{M}$ should be symmetric: $M_{i,j}=M_{j,i}$.
% \begin{itemize}
% 	\setlength{\itemsep}{3pt}
% 	\setlength{\parsep}{3pt}
% 	\setlength{\parskip}{3pt}

% 	\item If an object $\bm{o}_i$ has one attribute $\bm{a}_j$, we set $M_{i,j+N_o}=1$, \eg $\bm{o}_1$ ($\bm{u}_1$) and $\bm{a}_1$ ($\bm{u}_4$) in Fig.~\ref{fig:encoding} are connected and $M_{1,4}=1$.
% 	\vspace{-2pt}
% 	\item If one relationship $\bm{r}_k$ connects with two objects $\bm{o}_i$ and $\bm{o}_j$, we set $M_{i,k+N_o+N_a}=M_{j,k+N_o+N_a}=1$, \eg $\bm{r}_1$ ($\bm{u}_5$) connects $\bm{o}_1$ ($\bm{u}_1$) and $\bm{o}_2$ ($\bm{u}_2$) and $M_{5,1}=M_{5,2}=1$.
% 	\vspace{-2pt}
% 	\item All the object nodes are connected with each other since they are visual features that their contexts play a key role in captioning~\cite{herdade2019image}. Thus $\forall i,j\leq N_o, M_{i,j}=1$.
% 	\vspace{-2pt}
%     \item Since the scene graph is an undirected graph, $\bm{M}$ should be symmetric: $M_{i,j}=M_{j,i}$.

% \end{itemize}

After getting $\bm{U}$ and $\bm{M}$, we can revise the Transformer encoder to get our MHA-GNN. Specifically, we use $\bm{U}$ as the input of the encoder defined in Eq.\eqref{equ:encoder} and revise the \textbf{Att} operation in Eq.~\eqref{equ:multi-head} as the following \textbf{Masked Att} operation:
\begin{equation}
\small
    \bm{A}_i=\text{Softmax}(M \odot \frac{\bm{Q}\bm{W}_i^Q(\bm{K}\bm{W}_i^K)^T}{\sqrt{d}} ), \\
\end{equation}
where $\odot$ denotes the element-wise product.
In this way, the graph operation is defined by MHA. Specifically, for each node embedding, it is updated by weighted summing its neighbour embeddings, where the weights are from the attention heads $\bm{A}_i$ calculated by the \textbf{Att} operation in Eq.~\eqref{equ:multi-head}. During weighted summing, the binary matrix control whether two nodes are connected or not. Note that the edge type is implicitly embedded in \textbf{Att} operation due to the added node type embedding. For example, after adding node type embeddings $\bm{e}_o$ and $\bm{e}_r$ to the object and relation embeddings $\bm{o}$ and $\bm{r}$, respectively, the inner-product becomes\footnote{For convenience, we omit the trainable matrices $\bm{W}^Q,\bm{W}^K$ in Eq.~\eqref{equ:multi-head} in this inner-product operation.}:
\begin{equation}
\label{equ:inner_prod}
\small
    (\bm{o}+\bm{e}_o)^T(\bm{r}+\bm{e}_r) = \bm{o}^T\bm{r}+\bm{e}_o^T\bm{r}+\bm{e}_r^T\bm{o}+\bm{e}_o^T\bm{e}_r,
\end{equation}
where the right three terms are affected by the node type embedding. Thus, when the edge type changes (\eg the object-relation edge changes to object-attribute edge), the corresponding node type embeddings also change (\eg $\bm{e}_r$ changes to $\bm{e}_a$), which means Eq.~\eqref{equ:inner_prod} encodes the knowledge of edge types into the embeddings. 

By stacking more such layers, the receptive field is increasing and thus each node can be updated by aggregating more neighbour embeddings, which naturally follows the design principle of GNN~\cite{battaglia2018relational}. The output graph embedding set $\bm{G}$ are input to the decoder for captioning.

\subsection{MOE-decoder}
\begin{figure}[t]
  \centering
  \includegraphics[width=7.5cm]{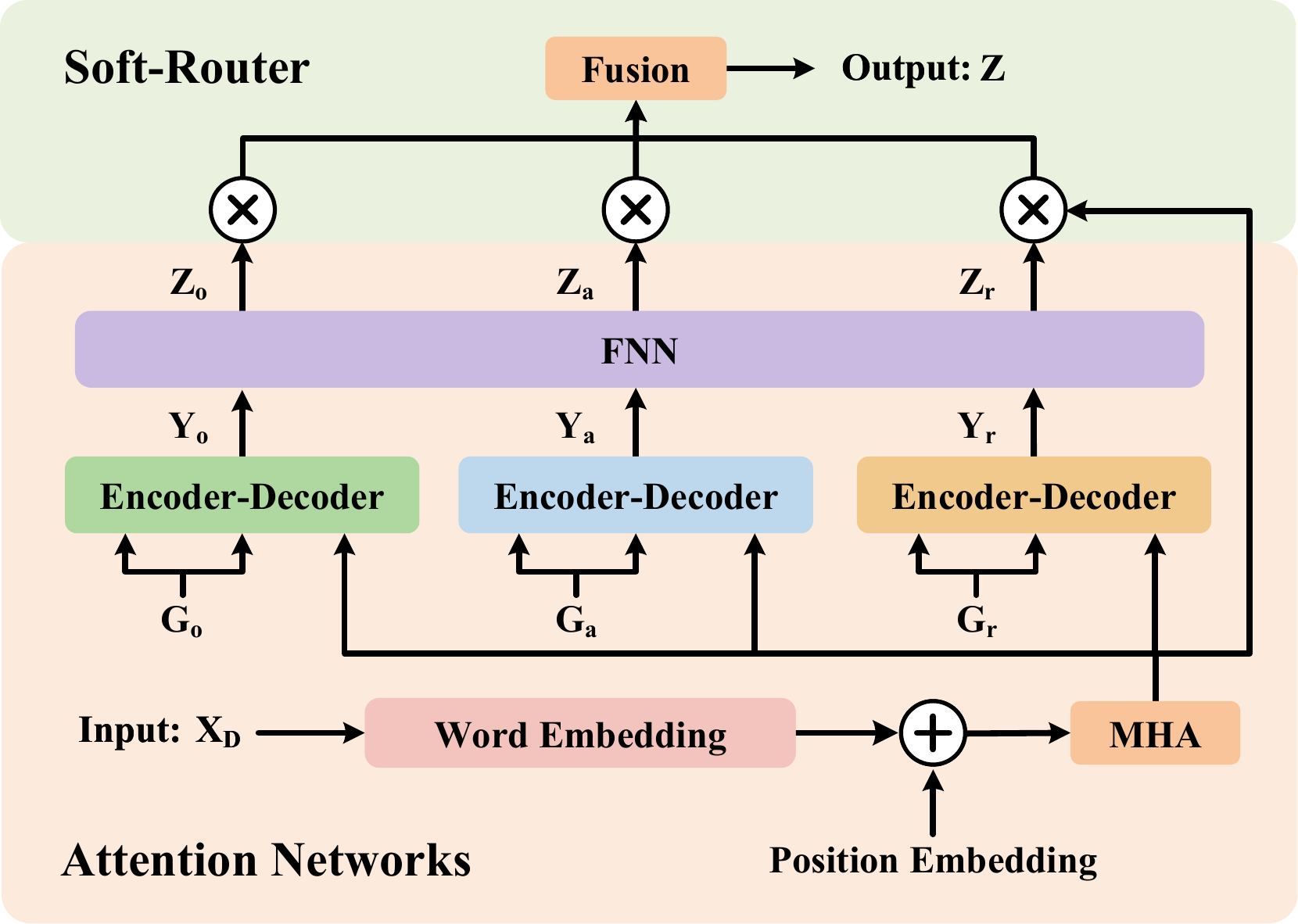}
  \caption{The sketch of the MOE-decoder, we use different colours to denote different experts: green/blue/yellow correspond to object/attribute/relation experts.}
  \label{fig:decoding}
\end{figure}
\label{sec:TFSGC_decoder}
As mentioned before, a caption contains different kinds of words for describing diverse visual patterns, \eg nouns/adjectives/verbs for objects/attributes/relations~\cite{yang2019learning}, which suggests that different experts should be used to address diverse visual knowledge for generating the corresponding words. Motivated by this idea, we design an MOE-based~\cite{jacobs1991adaptive,du2022glam} language decoder to discriminate diverse graph embeddings by setting three encoder-decoder attention layers as different experts. As shown in Fig.~\ref{fig:overall} (c), the graph embeddings $\bm{G}=\{ \bm{g}_1,...,\bm{g}_{N} \}$ output from the MHA-GNN can be naturally divided according to the original token types in the scene graph: object/attribute/relation sets $\bm{G}_o=\{\bm{g}_1,...,\bm{g}_{N_o}\}$/ $\bm{G}_a=\{\bm{g}_{N_o+1},...,\bm{g}_{N_o+N_a}\}$/$\bm{G}_r=\{\bm{g}_{N_o+N_a+1},...,\bm{g}_{N}\}$. Then we only need to input them into the corresponding experts for discriminating them. Fig.~\ref{fig:decoding} sketches the designed MOE-based decoder, which is got by revising the decoder defined in Eq.~\eqref{equ:decoder} as:
\begin{equation} \label{equ:decoder_TFSGC}
\small
\begin{aligned}
 \textbf{Input:} \quad  &\bm{X}_D,\bm{G}_o,\bm{G}_a,\bm{G}_r \\
 \textbf{SA:} \quad  &\bm{X}=\textbf{MHA}(\bm{Q}=\bm{K}=\bm{V}=\bm{X}_D), \\
 \textbf{EXP$_{O}$:} \quad  &\bm{Y}_o=\textbf{MHA}(\bm{Q}=\bm{X},\bm{K}=\bm{V}=\bm{G}_o), \\
 \textbf{EXP$_{A}$:} \quad  &\bm{Y}_a=\textbf{MHA}(\bm{Q}=\bm{X},\bm{K}=\bm{V}=\bm{G}_a), \\
 \textbf{EXP$_{R}$:} \quad  &\bm{Y}_r=\textbf{MHA}(\bm{Q}=\bm{X},\bm{K}=\bm{V}=\bm{G}_r), \\
 \textbf{FFN:}  \quad  &\bm{Z}_o,\bm{Z}_a,\bm{Z}_r=\textbf{FFN}(\bm{Y}_o,\bm{Y}_a,\bm{Y}_r), \\
 \textbf{SR:}  \quad  &\bm{Z}=\textbf{SR}(\bm{Z}_o,\bm{Z}_a,\bm{Z}_r,\bm{X})
\end{aligned}
\end{equation}
where \textbf{EXP$_{O}$}, \textbf{EXP$_{A}$}, and \textbf{EXP$_{R}$} denote three different experts (encoder-decoder attentions) used to address object, attribute, and relation embeddings, respectively. They have the same structure while with different parameters. 

Note that the input $\bm{X}_D$ is the word embeddings of the partially generated captions and at $t$-th step, $\bm{X}_D=\{\bm{x}_D^1,...,\bm{x}_D^t\}$. Then all the $\bm{X}$/$\bm{Z}_o$/$\bm{Z}_a$/$\bm{Z}_r$ also contain $t$ elements, \eg $\bm{Z}_o=\{\bm{z}_o^1,...,\bm{z}_o^t\}$. Soft Router (\textbf{SR}) calculates an ensemble embedding $z$ at each time step to construct the embedding set $\bm{Z}=\{\bm{z}^1,...,\bm{z}^t\}$. Specifically, for each element $\bm{x}$/$\bm{z}_o$/$\bm{z}_a$/$\bm{z}_r$ in $\bm{X}$/$\bm{Z}_o$/$\bm{Z}_a$/$\bm{Z}_r$, a corresponding output $\bm{z}$ can be got\footnote{For convenience, we remove the superscript representing for different time steps of each embedding.}:
\begin{equation} \label{equ:selector}
\small
\begin{aligned}
 \textbf{Input:} \quad  &\bm{x},\bm{z}_o,\bm{z}_a,\bm{z}_r ,\\
 \textbf{ATT:} \quad  &
 \begin{aligned}
     \bm{\alpha}&=\{\alpha_o,\alpha_a,\alpha_r\}\\
     &=\text{Softmax}( \{\bm{x}^T\bm{z}_o,\bm{x}^T\bm{z}_a,\bm{x}^T\bm{z}_r\} )
 \end{aligned} \\
 \textbf{Output:} \quad  &\bm{z}= \alpha_o\bm{z}_o+\alpha_a\bm{z}_a+\alpha_r\bm{z}_r,\\
\end{aligned}
\end{equation}
where \textbf{ATT} operation calculates the soft routing weights, since $\bm{x}$ accumulates the context knowledge of the partially generated caption, it can help judge which kind of word should be generated at the next step. For example, if the last word of this partially generated caption is an adjective ``black'', the next word is more like to be a noun and thus $\alpha_o$ should be a large value for using more object embeddings instead of the other embeddings.

\section{Experiments}
\subsection{Datasets, Metrics, and Implementation Details}
\noindent\textbf{Datasets.}
\noindent\textbf{MSCOCO.}
We use MSCOCO~\cite{lin2014microsoft} to validate our TFSGC. This dataset has 123,287 images and each one is labeled with 5 captions. We use two splits in the experiments: the offline Karpathy split (113,287/5,000/5,000 train/val/test images) and the Official online split (82,783/40,504/40,775 train/val/test images). 
% We lowercase the words and trim the sentences to a maximum of 16 words. The words which appear less than 5 times are deleted and remain a vocabulary with $10, 369$ words.

\noindent\textbf{Visual Genome~\cite{krishna2017visual}} provides scene graph annotations for training the scene graph parser. We follow~\cite{yang2020auto} to filter the noisy dataset (\eg lots of labels only appear a few times in the dataset) by removing the attribute/relation labels appearing less than 2000 times and use the remained 103/64 attribute/relation labels to train the attribute/relation classifiers.

\noindent\textbf{Implementation Details.}
% In the experiments, we use two kinds of visual features, which are regional-based ones extracted from the ResNet~\cite{ren2015faster,anderson2018bottom} and grid-based ones extracted from the visual Transformer~\cite{liu2021swin}. For both visual features, we set the batch size to 10 and use Adam~\cite{kingma2014adam} as the optimizer. For regional/grid features, We sequentially use cross-entropy loss (Eq.~\eqref{equ:equ_celoss}) and RL-based reward (Eq.~\eqref{equ:equ_rlloss}) to train the models 20/30 and 30/30 epochs. For regional/grid features, the learning rate used in the cross-entropy stage is initialized as $5e^{-4}$/$2e^{-5}$ and both decayed by 0.8 every 5 epochs, the learning rate used in the RL-reward stage is reset to $2e^{-5}$/$5e^{-6}$ and both decayed by 0.8 every 5 epochs. During inference, we use beam search where the beam size is 5. We evaluate the captions by CIDEr-D~\cite{vedantam2015cider}, BLEU~\cite{papineni2002bleu}, METEOR\cite{banerjee2005meteor}, ROUGE~\cite{lin2004rouge} and SPICE~\cite{anderson2016spice}. 
In the experiments, we use three kinds of visual features to exhaustively compare to the other SOTA models, which are BUTD~\cite{anderson2018bottom}, Patch~\cite{liu2021swin}, and VinVL~\cite{zhang2021vinvl}. During training and inference, for BUTD/Patch/VinVL, we respectively follow ~\cite{yang2020auto} and VinVL’s official parser\footnote{\url{https://github.com/microsoft/scene_graph_benchmark}} to parse SGs, where the latter is more powerful. For all the visual features, we set the batch size to 20 and use Adam~\cite{kingma2014adam} as the optimizer. For BUTD/Patch/VinVL features, We sequentially use cross-entropy loss (Eq.~\eqref{equ:equ_celoss}) and RL-based reward (Eq.~\eqref{equ:equ_rlloss}) to train the models 20/20/30 and 30/30/30 epochs. For BUTD/Patch/VinVL features, the learning rate used in the cross-entropy stage is initialized as $5e^{-4}$/$5e^{-4}$/$2e^{-5}$ and both decayed by 0.8 every 5 epochs, the learning rate used in the RL-reward stage is reset to $5e^{-5}$/$2e^{-5}$/$5e^{-6}$ and both decayed by 0.8 every 5 epochs. During inference, we use beam search where the beam size is 5. We evaluate the captions by CIDEr-D~\cite{vedantam2015cider}, BLEU~\cite{papineni2002bleu}, METEOR\cite{banerjee2005meteor}, ROUGE~\cite{lin2004rouge} and SPICE~\cite{anderson2016spice}. 

\begin{figure*}[htb]
  \centering
  \includegraphics[width=16cm]{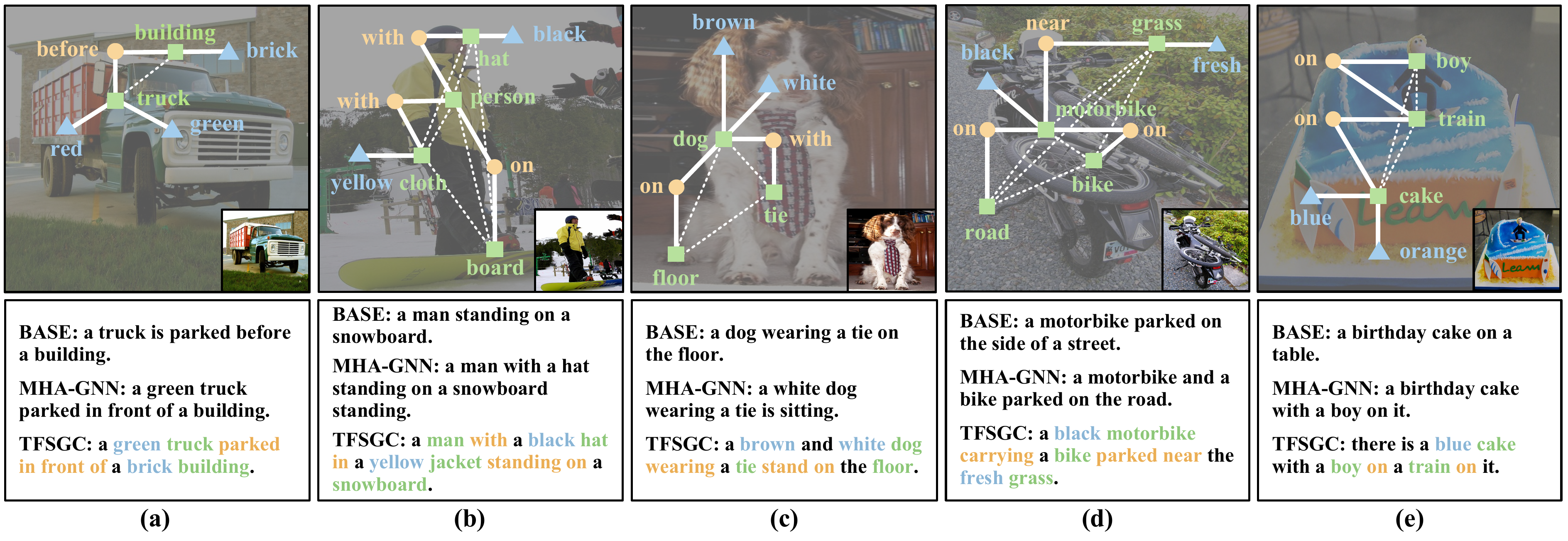}
  \caption{The captions generated by \textbf{BASE}, \textbf{MHA-GNN}, and \textbf{TFSGC}. It can be found that TFSGC generates more descriptive captions, \eg it describes more attributes like ``green truck'' and ``brick building'' in (a), or using more fine-grained nouns like ``jacket'' in (b). Diverse colors in TFSGC denote these words use more knowledge from different experts: green/blue/orange corresponds to object/attribute/relation expert, which is got by checking which one of $\alpha_o$/$\alpha_a$/$\alpha_r$ in Eq~\eqref{equ:selector} is the largest.} \label{fig:fig_example}
  % \vspace{-0.2in}
\end{figure*}

\subsection{Ablation Studies}
% \noindent\textbf{Comparison Models}. 
To confirm the effectiveness of the proposed MHA-GNN and MOE-decoder, we deploy exhaustive ablations as follows. Note that we use BUTD feature in this section.
\textbf{BASE}: We directly use the classic Transformer architecture. 
\textbf{SG}: We incorporate the scene graphs into the Transformer by using the node embeddings without any graph operations.
\textbf{MLP-GNN}: We apply MLP-based Graph Neural Network~\cite{xu2018how} for embedding the scene graphs.
\textbf{MHA-GNN w/o $\bm{e}$}: We apply the proposed MHA-GNN while do not use node type embedding. 
\textbf{MHA-GNN}: We apply the proposed MHA-GNN and keep the decoder unchanged as BASE. 
\textbf{GNN-FC}:We remove the binary mask matrix $M$ from TFSGC.
\textbf{MOE}: We use the proposed MOE-decoder and do not use GNN but input the original node embeddings into the decoder.
\textbf{TFSGC}: We apply the integral TFSGC. 

\begin{table}[t]
\begin{center}
%ROUGE-L
\scalebox{0.78}{
\begin{tabular}{l  c c c c c}
		\hline
		   Models   & B@4 & M & R &   C & S\\ \hline
           BASE & $38.4$ & $28.5$ & $58.1$ & $128.7$ & $22.0$ \\
           SG & $38.5$ & $28.6$ & $58.1$ & $129.0$ & $22.2$ \\
           MLP-GNN & $38.9$ & $28.8$ & $58.4$ & $129.5$ & $22.4$ \\
           MHA-GNN w/o $\bm{e}$ & $39.1$ & $28.9$ & $58.5$ & $130.1$ & $22.4$ \\
           MHA-GNN & $39.5$ & $29.2$ & $58.9$ & $130.9$ & $22.8$ \\
           GNN-FC & $39.2$ & $29.0$ & $58.3$ & $130.5$ & $22.3$ \\
           MOE & $39.2$ & $28.8$ & $58.5$ & $130.1$ & $22.5$ \\
           TFSGC   &$\bm{39.8}$ & $\bm{29.6}$ & $\bm{59.3}$ & $\bm{132.3}$ & $\bm{23.4}$\\
            \hline
\end{tabular}    
}
\caption{The performances of various ablation models. The metrics: B@N, M, R, C, and S denote BLEU@N, METEOR, ROUGE-L, CIDEr-D, and SPICE.}
\label{table:tab_abla}
\end{center}
% \vspace{-0.2in}
\end{table}

\begin{table}[t]
\begin{center}
\scalebox{0.75}{
\begin{tabular}{l c c c c c}
		\hline
		   Models   & nouns & adjectives & verbs & prepositions \\ \hline
           BASE & $43.8$ & $12.7$ & $20.3$ & $40.1$ \\ 
           SG & $44.2$ & $13.5$ & $20.9$ & $40.8$ \\ 
           MLP-GNN & $45.4$ & $14.8$ & $21.8$ & $41.6$  \\
           MHA-GNN w/o $\bm{e}$ & $48.8$ & $16.4$ & $24.3$ & $44.3$  \\
           MHA-GNN & $49.6$ & $17.0$ & $24.7$ & $44.6$ \\
           MOE & $48.4$ & $16.6$ & $24.4$ & $44.0$  \\
           TFSGC & $\bm{52.8}$ & $\bm{18.2}$ & $\bm{25.8}$ & $\bm{45.8}$ \\ \hline 
\end{tabular}
}
\caption{The recalls (\%) of five part-of-speech words.}
\label{tab:recall}
\end{center}
\vspace{-0.2in}
\end{table}

% \noindent\textbf{Analyses of Similarity Metrics}. 
Table~\ref{table:tab_abla} compares the similarity metrics of the ablation models. Firstly, we can find that the integral TFSGC achieves the highest scores, which confirms its effectiveness. Next, we respectively compare the ablation models to validate the effectiveness of the proposed MHA-GNN and MOE-decoder. By comparing MLP-GNN, SG, and BASE, it is easy to see that using GNN can gain more profits than only using node embeddings. Furthermore, we can find that MHA-GNN has higher CIDEr than MLP-GNN, which suggests that designing GNN by MHA is more powerful than by MLP in the Transformer architecture. Next, to see whether discriminating the graph embeddings is beneficial or not, we can compare MHA-GNN with MHA-GNN w/o $\bm{e}$ and find that using node type embedding performs better. From the comparison between GNN-FC and TFSGC, it can be seen that when removing $M$, the graph becomes a fully-connected graph, which introduces more noises. Also, it can be seen that MOE and TFSGC respectively achieve better performances than SG and MHA-GNN, which validates the effectiveness of the MOE-decoder. 

% \noindent\textbf{Analyses of Descriptiveness}.  
Besides evaluating these ablation models by similarities, in Table~\ref{tab:recall}, we calculate the recalls of the words with different POS to evaluate the descriptiveness. Table~\ref{tab:recall} shows that the captions generated from TFSGC have the highest recalls, suggesting that TFSGC generates the most descriptive captions. Also, we can find that both the proposed MHA-GNN (MHA-GNN vs. MLP-GNN) and MOE-based decoder (MOE vs. SG) can boost the recalls, suggesting both of them improve the descriptiveness. Moreover, We use Stanford Parser to get the POS to train the route weights by the cross-entropy loss, then the CIDEr of TFSGC boosts from 132.3 to 132.9, suggesting the advantage of using POS knowledge. Fig.~\ref{fig:fig_example} shows 4 examples of the captions generated from diverse models, where we can see that TFSGC generates more descriptive captions. BASE generates less descriptive captions since it does not use scene graphs and thus loses semantic knowledge compared with SG. MHA-GNN does not use the soft router and thus can not select the most suitable experts for generating corresponding words, which may lose certain details. Also, we show that which expert is more responsible for the words with different POS, \eg in Fig.~\ref{fig:fig_example} (b), the adjective ``yellow'' is generated by using more knowledge from the attribute expert.

\begin{table*}[t]
\begin{center}
\setlength{\tabcolsep}{3pt}
\vspace{-0.2in}
%ROUGE-L
% \scalebox{0.8}{
\begin{tabular}{lccccccccccc}
\hline
\multirow{2}{*}{Models} & \multicolumn{5}{c}{Cross-Entroy Loss} & &\multicolumn{5}{c}{CIDEr optimization} \\
\cmidrule(r){2-6}  \cmidrule(r){8-12}
                        & \multicolumn{1}{c}{B@4} & \multicolumn{1}{c}{M} & \multicolumn{1}{c}{R} & \multicolumn{1}{c}{C} & \multicolumn{1}{c}{S} & & \multicolumn{1}{c}{B@4} & \multicolumn{1}{c}{M} & \multicolumn{1}{c}{R} & \multicolumn{1}{c}{C} & \multicolumn{1}{c}{S} \\
\hline
           {$^{\color{red!80}{{\text{\textbf{BUTD feature}}}}}$} & & & & & & & & & \\[-6pt] 
           ETA   &$37.1$&$28.2$&$57.1$&$117.9$ &$21.4$&& $39.3$ & $28.8$ & $58.9$ & $126.6$ & $22.7$ \\ 
           ORT &$35.5$&$28.0$&$56.6$&$115.4$&$21.2$&  & $38.6$ & $28.7$ & $58.4$ & $128.3$ & $22.6$ \\
           AoANet  &$37.2$&$28.4$&$57.5$&$119.8$&$21.4$&   & $38.9$ & $29.2$ & $58.8$ & $129.8$ & $22.4$ \\
           $\mathcal{M}^2$ Transformer   &-&-&-&-&-&   & $39.1$ & $29.2$ & $58.6$ & $131.2$ & $22.6$ \\ 
            CATT &$37.3$&$28.5$&$57.4$&$119.0$&$21.5$& &$39.4$&$29.3$& $58.9$&$131.7$&$22.8$\\ 
           APN &-&-&-&-&-& & $39.6$ & $29.2$ & $59.1$ & $131.8$ & $23.0$\\
           DLCT&-&-&-&-&-&   & $\bm{39.8}$&$29.5$&$59.1$&$\bm{133.8}$&$23.0$\\
           TFSGC  &$\bm{38.1}$ & $\bm{28.6}$ & $\bm{57.7}$ & $\bm{120.2}$ & $\bm{21.9}$ &&$\bm{39.8}$ & $\bm{29.6}$ & $\bm{59.3}$ & $132.3$ & $\bm{23.4}$\\
           \hline
           {$^{\color{red!80}{{\text{\textbf{BUTD feature \& Larger Batch Size}}}}}$} & & & & & & & & & \\[-6pt] 
           X-Transformer&$38.2$&$\bm{28.8}$&$\bm{58.0}$&$122.0$&$21.9$& &$39.7$&$29.5$& $59.2$&$132.8$&$23.2$\\
           TFSGC$^*$ & $\bm{38.4}$ & $\bm{28.8}$ & $57.8$ & $\bm{122.3}$ & $\bm{22.1}$& & $\bm{39.9}$ & $\bm{29.8}$ & $\bm{59.4}$ & $\bm{133.0}$ & $\bm{23.4}$\\
           \hline
           {$^{\color{red!80}{{\text{\textbf{Large Visual-language model pretraining}}}}}$} & & & & & & & & & \\[-6pt]
           \rowcolor{gray!20}
           RSTNet& -&- & - & - & -&&$40.1$&$28.9$&$59.5$&$135.6$&$23.3$\\
            \rowcolor{gray!20}
            LEMON$_{base}$&$40.3$& $30.2$&-&$133.3$&$ 23.3$&&$41.6$&$30.1$&-&$142.7$& $25.1$\\
            \rowcolor{gray!20}
            LEMON$_{huge}$&$41.5$& $30.8$&-&$139.1$&$ 24.1$&&$42.6$&$31.4$&-&$145.5$& $25.5$\\
           \hline
           {$^{\color{red!80}{{\text{\textbf{Patch feature}}}}}$} & & & & & & & & & \\[-6pt] 
           PureT&-&-&-&-&- && $40.9$ &$ \bm{30.2}$ & $\bm{60.1}$ & $138.2$ & $24.2$  \\
           ViTCAP$_{small}$ &$35.7$& $28.8$& $57.6$&$121.8$&$ 22.1$&&$40.1$&$29.4$&$59.4$&$133.1$& $23.0$\\
           % ViTCAP-large &$36.3$& $29.3$& $ 58.1$&$125.2$&$ 22.6$&&$41.2$&$30.1$&$60.1$&$138.1$& $24.1$\\
           TFSGC & $\bm{38.8}$ & $\bm{29.4}$ & $\bm{58.2}$ & $\bm{122.2}$ & $\bm{22.3}$ && $\bm{41.4}$ & $30.1$ & $\bm{60.1}$ & $\bm{138.6}$ & $\bm{24.4}$\\
           \hline
           {$^{\color{red!80}{{\text{\textbf{VinVL feature}}}}}$} & & & & & & & & & \\[-6pt] 
           VinVL(Transformer)  &35.4&28.6&57.5&121.5&$21.3$ && $40.6$ &$30.0$ & $59.8$ & $137.3$ & $23.7$  \\
           TFSGC  & $\bm{38.5}$ & $\bm{29.2}$ & $\bm{58.8}$ & $\bm{122.7}$ & $\bm{22.4}$ && $\bm{41.7}$ & $\bm{30.5}$ & $\bm{60.4}$ & $\bm{139.5}$ & $\bm{24.6}$\\
           % \hline
           % {$^{\color{red!80}{{\text{\textbf{Visual-language BERT pretraining}}}}}$} & & & & & & & & & \\[-6pt] 
           % RSTNet & -&- & - & - & -&&$40.1$&$28.9$&$59.5$&$135.6$&$23.3$\\
          
           \hline
\end{tabular}
% }
\caption{The performances of SOTA methods on MS-COCO Karpathy split. All models used are single models.}
\label{table:tab_sota}
\end{center}
% \vspace{-0.1in}
\end{table*}

\begin{table}[t]
\begin{center}
\scalebox{0.55}{
\begin{tabular}{l c c c c c c c c}
		\hline
		\multirow{2}*{Models}   & \multicolumn{2}{c}{B@4} & \multicolumn{2}{c}{M} & \multicolumn{2}{c}{R} &  \multicolumn{2}{c}{C} \\ \cmidrule(r){2-3} \cmidrule(r){4-5} \cmidrule(r){6-7} \cmidrule(r){8-9} 
		   & c5 & c40 & c5 & c40 & c5 & c40 & c5 & c40\\
		    \hline
           Up-Down & $36.9$  & $68.5$  & $27.6$  & $36.7$  & $57.1$  & $72.4$  & $117.9$  & $120.5$ \\ 
           SGAE & $37.8$  & $68.7$  & $28.1$  & $37.0$  & $58.2$  & $73.1$  & $122.7$  & $125.5$ \\
           ETA & $38.9$ & $70.2$ & $28.6$ & $38.0$ & $58.6$  & $73.9$  & $122.1$  & $124.4$ \\ 
           APN & $38.9$ & $70.2$ & $28.8$ & $38.0$ & $58.7$ & $73.7$ & $126.3$ & $127.6$\\
           NG-SAN  &  38.8 & 70.2 & 29.0 & 38.4 & 58.7 & 74.0 & 126.3 & 128.6 \\
           TFSGC$^S$    & $\bm{39.0}$ & $\bm{70.9}$ & $\bm{29.1}$ & $\bm{38.4}$ & $\bm{58.9}$ & $\bm{74.4}$ & $\bm{127.2}$ & $\mathbf{129.8}$\\ \hline
           AoANet & $39.4$ & $71.2$ & $29.1$ & $38.5$ & $58.9$ & $74.5$ & $126.9$ & $129.6$ \\
           X-Transformer & $\bm{39.9}$  & 71.8  & 29.5  & 39.0  & 59.3  & 74.9  & 129.3  & 131.4 \\
           $\mathcal{M}^2$ Transformer& 39.7 & $\bm{72.8}$ & 29.4 & 39.0 & 59.2 & 74.8 & 129.3 & 132.1 \\
           TFSGC$^E$    & $39.7$ & $72.6$ & $\bm{29.6}$ & $\bm{39.2}$ & $\bm{59.5}$ & $\bm{75.1}$ & $\bm{129.6}$ & $\bm{133.1}$\\ \hline
\end{tabular}
}
\caption{The scores on the MS-COCO online test server. 
 $^S$ indicates single model, $^E$ indicates ensemble model.}
\label{table:tab_cap_on}
\end{center}
\vspace{-0.2in}
\end{table}

\subsection{Comparisons with SOTA}
Recently, various SOTA captioning models with diverse settings have been proposed, including using different language decoders (LSTM, GRU, and Transformer), different visual features, and whether distilling knowledge from large-scale pre-training models (CLIP~\cite{radford2021learning} or VInVL~\cite{zhang2021vinvl}). In fairness, we compare our TFSGC with the models that also use Transformer-based decoder by three features: BUTD~\cite{anderson2018bottom}, Patch~\cite{liu2021swin}, and VinVL~\cite{zhang2021vinvl}. 
% Note that we do not compare with extra large scale models trained by millions of image-text pairs. 

We compare with the following SOTA models: \textbf{ETA}~\cite{Li_2019_ICCV}, \textbf{ORT}~\cite{herdade2019image}, \textbf{AoANet}~\cite{huang2019attention},
$\mathcal{M}^2$ \textbf{Transformer}~\cite{cornia2020meshed},
\textbf{CATT}~\cite{yang2021causal},
\textbf{APN}~\cite{yang2021auto}, 
\textbf{DLCT}~\cite{luo2021dual}, 
\textbf{X-Transformer}~\cite{pan2020x},
\textbf{Up-Down}~\cite{anderson2018bottom},
\textbf{SGAE}~\cite{yang2019auto},
\textbf{NG-SAN}~\cite{guo2020normalized},
\textbf{PureT}~\cite{2022End}, and \textbf{ViTCAP}~\cite{fang2022injecting}. Specifically,
ETA and ORT are preliminary Transformer-based models; AoANet and X-Transformer design stronger attention mechanisms; CATT and $\mathcal{M}^2$ Transformer introduce additional memory networks; and APN exploits the hidden tree structures. We set the training batch size of TFSGC$^*$ to 50 as X-Transformer for fairly comparing. These methods use BUTD features, while ViTCAP and PureT use Patch features. RSTNet and ViTCAP distill knowledge from pre-trained vision-language BERTs. Note that in VinVL~\cite{zhang2021vinvl}, they use OSCAR~\cite{li2020oscar} as the captioner, while OSCAR is trained on 6.5 million image-text pairs. To fairly compare, we input VinVL feature into the classic Transformer to generate the captions, which is denoted as \textbf{VinVL(Transformer)}. Note that we do not compare with extra large scale models trained by millions of image-text pairs or with excessively large parameter sizes, such as \textbf{RSTNet}~\cite{zhou2020unified} and \textbf{LEMON}~\cite{hu2022scaling}. All the results are shown in Table~\ref{table:tab_sota}.

% From Table~\ref{table:tab_sota} we can find that TFSGC achieves the highest  CIDEr-D scores in different settings,
From Table~\ref{table:tab_sota} we can find that TFSGC almostly achieves the highest scores in different settings, \ie achieving 132.3, 138.6, 139.5 CIDEr scores when using BUTD, Patch, and VinVL features, respectively. DLCT’s visual features extractor ~\cite{jiang2020defense} is stronger than TFSGC(BUTD) ~\cite{anderson2018bottom} and thus DLCT is a little better than TFSGC(BUTD). Among these compared methods, although other SOTAs do not use scene graphs, they usually have some other training burdens. For example, APN and X-Linear apply more complex attention mechanisms and it requires more computation resource for well-training them, while our TFSGC only apply the simplest attention operation. Moreover, as detailed in Sec 4.1 of ViTCAP, it applies much more training data (9.9M image-text pairs from 4 datasets including VG) to pre-train a concept network to get more powerful discrete tags, while we only use one dataset VG to get scene graphs and achieve better performance, which suggests that connecting discrete tags into a graph is a useful strategy if the discrete tags is not very powerful. Moreover, TFSGC achieves comparable performances with LEMON~\cite{hu2022scaling} which uses 200 million image-text pairs and uses 12/24-layer, 768/1024-hidden units Transformers(base/huge).
To sum up, the advantage of TFSGC is that it can effectively embed and discriminate the semantic knowledge from scene graphs to balance the (usually more) burden of using more training data or of training more complex networks.

We also submit the single model TFSGC$^S$ and 4-ensembled model TFSGC$^E$ trained by regional-based features into the online server for testing, where the results are shown in Table~\ref{table:tab_cap_on}. From this table, we can discover that both TFSGC$^S$ and TFSGC$^E$ have the highest CIDEr-D scores, which further confirm the effectiveness of the proposed TFSGC.

\section{Conclusion}
We proposed to transform the scene graph into captions (TFSGC) by a simple and homogeneous network. Specifically, we use MHA to design the GNN by linearizing the scene graph and remedying the lost topological knowledge with a binary mask matrix. Furthermore, we add learnable type embedding and design an MOE-based decoder to distinguish node embeddings for more descriptive captions. At last, we compared TFSGC with various SOTA models and demonstrated that our model can achieve comparable performances to some strong benchmarks.

\section*{Limitations}
There are two major limitations of the proposed TFSGC. The first one is that the effectiveness of TFSGC depends on the quality of the scene graph. Since MSCOCO does not have SG annotations, we evaluate the parsers in Visual Genome: for BUTD/Patch and VinVL, the recall@50 of relation/attribute are respectively 65.2/68.4 and 73.4/76.6. We use VinVL's SGs in TFSGC(BUTD) and CIDEr improves from 132.3 to 133.1, suggesting better SGs are beneficial.  If the scene graph quality is poor, then TFSGC will not achieve good performance. When an incorrect node in the scene graph, it also affects the output of the caption. \eg in Fig.~\ref{fig:fig_example} (e), the correct object label should be "surfboard" instead of "train".
% In this paper, we use Visual Genome, which contains abundant and useful scene graph annotations for parsing effective scene graphs and thus TFSGC is powerful.
In this paper, we use Visual Genome, which contains abundant and useful scene graph annotations for parsing effective scene graphs, but current performance is not the best, and we will improve the scene graph parser based on the latest scene graph parsing methods in the future.

The second limitation of TFSGC is that if the visual features contain 
abundant attribute or relation knowledge, then the improvement of TFSGC compared with the classic Transformer will be weakened. For example, compared with the BUTD feature case where the relative improvement of CIDEr-D is 3.6 (TFSGC-BASE in Table~\ref{table:tab_abla}), the VinVL feature is more powerful since it is trained by much more data samples with more semantic labels, thus the relative improvement is lower, which is 2.2 (TFSGC-VinVL(Transformer) in Table~\ref{table:tab_sota}).

\section*{Acknowledgements}
This work is supported by National Key R\&D Program of China (2018AAA0100104, 2018AAA0100100), National Science Foundation of China (62206048), Natural Science Foundation of Jiangsu Province (BK20220819, BK20211164), and Young Elite Scientists Sponsorship Program of Jiangsu Association for Science and Technology Tj-2022-027.

% \clearpage
\bibliography{anthology,acl2023}
\bibliographystyle{acl_natbib}

\end{document}